\documentclass[letterpaper, 10 pt, conference]{ieeeconf}  

\IEEEoverridecommandlockouts                              

\usepackage{graphicx}
\usepackage{dblfloatfix}
\usepackage{url}
\usepackage{hyperref}

\title{\LARGE \bf Alternative Interfaces for Human-initiated Natural Language Communication and Robot-initiated Haptic Feedback: Towards Better Situational Awareness in Human-Robot Collaboration}

\author{Callum Bennie*, Bridget Casey*, Cécile Paris, Dana Kulić, Brendan Tidd, Nicholas Lawrance, \\ Alex Pitt, Fletcher Talbot,  Jason Williams, David Howard, Pavan Sikka, and Hashini Senaratne
\thanks{All authors are/were with Robotics and Autonomous Systems group and/or Collaborative Intelligence Future Science Platform (CINTEL FSP) of CSIRO Data61, Australia. Dana Kulic is also with Monash University, Australia. This paper covers work carried out during two vacation student projects funded by CSIRO Data61 Robotgals Scholarship. *Callum Bennie and Bridget Cassey equally contributed to this work, as the main student contributors of the natural language communication project and the haptic feedback project, respectively. We thank Annabelle Knott, Ze'ev Krischer and users engaged with preliminary evaluations for their valuable feedback.
Note: This paper was peer reviewed and published at \href{https://workshophri.github.io/OzCHI2023/}{Empowering People in Human-Robot Collaboration: Why, How, When, and for Whom} workshop at \href{http://www.ozchi.org/2023/index.html}{OzCHI 2023} conference.}
}

\begin{document}

\maketitle
\thispagestyle{empty}
\pagestyle{empty}

\begin{abstract}

This article presents an implementation of a natural-language speech interface and a haptic feedback interface that enables a human supervisor to provide guidance to, request information, and receive status updates from a Spot robot. We provide insights gained during preliminary user testing of the interface in a realistic robot exploration scenario.

\end{abstract}

\section{Introduction}

Traditionally, the application of robotics is largely focused on replacing human work in dull, dirty, and dangerous jobs~\cite{takayama2008beyond}. However, with more effective human-robot interaction technologies that enable humans and robots to work together on collaborative missions, the function of robots could extend to many more areas, such as personalized assistants~\cite{johnson2008collaborative} or human-robot teams for emergency response~\cite{ackerman2022robots}. 

Current interfaces that are designed for human-robot collaboration are mostly text and visual-based~\cite{berg2020review}. Robot supervision using these graphical user interfaces (GUIs) often requires expert knowledge and familiarity, creating a barrier to widespread adoption. Further, current graphical interfaces are complex and come along with a number of text and visual elements that the human team members need to continuously monitor. As a result, human team members often experience delays in detecting important status changes of robots~\cite{senaratne2023roman}.

As robots become increasingly prevalent in both industrial and domestic settings, the ability to communicate with them in a manner akin to human conversation could significantly enhance their usability and accessibility. Also, the ability to receive the selected status of robots through haptic feedback can be useful not to miss important information when busy with other duties or roles. 

To achieve these goals, we designed and implemented an interface that enables natural conversation with and haptic feedback from a Boston Dynamics Spot robot\footnote{\url{https://bostondynamics.com/products/spot}}, as alternative modalities to an existing GUI developed to perform disaster response and smart agriculture missions by a human-robot team ~\cite{chen2022multi}, to facilitate natural and intuitive communication with robots. The resulting multimodal interface is expected to reduce delays and failures associated with situational awareness and improve accessibility, thereby increasing team performance and broadening the potential user base.

\section{Human-initiated Natural Language Communication with Robots}

Recent advancements in Large Language Models (LLMs) such as GPT-3 and BERT support the rapid and low-cost development of flexible and natural communication systems compared to traditional Natural Language Processing (NLP) methods that use hand-crafted grammar rules and tailored models~\cite{min2023recent}. We integrated LLMs with speech recognition and synthesis technologies, expanding the possibilities for human-robot communication beyond the constraints of GUIs.

\subsection{Implementation of Speech Interface}

\subsubsection{System Overview} Our implementation focused on enabling a human team member to (1) request summaries of internal status from a Spot robot of the existing system (2) provide it with simple commands to execute, and (3) ask about the capabilities of its speech interface. To achieve this, we implemented a pipeline that utilised (1) OpenAI's Whisper model to transcribe user speech input to text\footnote{\url{https://platform.openai.com/docs/models/whisper}}, (2) OpenAI's ``text-davinci-003'' GPT module along with prompt engineering guidance developed by us to interpret user input and generate relevant text-based responses\footnote{\url{https://platform.openai.com/docs/models/gpt-3-5}}, and (3) Silero V3 module to perform text-to-speech synthesis\footnote{\url{https://github.com/snakers4/silero-models}}. 

\begin{figure*}[!htb]
\centering
\includegraphics[width=0.9\linewidth]{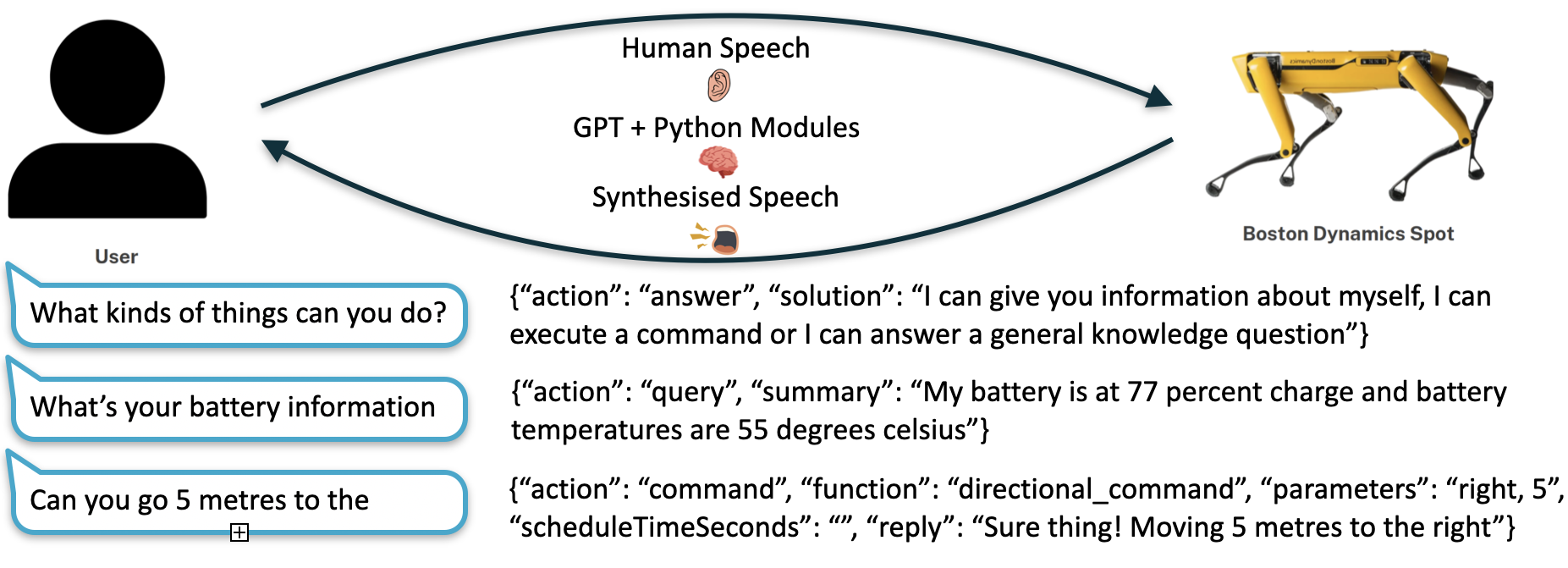}
\caption{System overview with sample user requests and GPT responses}
\label{fig:overview_examples}
\vspace{-4mm}
\end{figure*}

The outcome interface could be used by wearing a Bluetooth headset, which enables mobility while communicating with the robot. So, the human team member can use this interface while sitting at the GUI or performing a different task away from the GUI. Figure \ref{fig:overview_examples} illustrates the speech interface overview and example user inputs and responses from the GPT module (following a JSON format).


\subsubsection{Prompt creation}Our implementation utilized a variety of prompt engineering techniques. To establish context for the speech interface, we created a prompt for the GPT module explicitly stating that it is a robot assistant and that its responses should be helpful and accurate. This kind of prompt creation is in line with other attempts to condition performance for particular tasks~\cite{vemprala2023chatgpt}. 

Based on the retrieval augmented prompting (RAG) technique, we also added real-time updates of the robot's internal states into the prompt (i.e., retrieval-augmented prompting) to overcome shortcomings in LLM knowledge specific to the application, as done elsewhere\cite{cui2023chatlaw}. RAG process has three key parts: (1) Retrieval: Based on the prompt, retrieve relevant knowledge from a knowledge base, (2) Augmentation: Combine the retrieved information with the initial prompt, (3) Generate: pass the augmented prompt to a LLM, generating the final output. This facilitated the user to learn the robot's states, including executing tasks, error messages, battery information, and specific sensor readings. With the use of general GPT functionality, the user could engage in follow-up conversations about the robot's status. For instance, they could inquire about the implications of specific sensor levels or seek further details about a particular sensor, as a means to enhance the user's understanding and thereby control over the robot's operations.

Furthermore, we provided GPT with a command API in the prompt, detailing key parameters and function descriptions to enable the model to interpret user requests and execute simple commands. This technique is similar to current efforts to extend the capabilities of LLMs into task execution~\cite{schick2023toolformer}. With this, users could request the Spot robot to perform actions such as sitting, standing, moving specified distances in specified directions, recalling and returning to past locations, scheduling an action after a specified time duration, and operating onboard devices like lights. GPT extracted appropriate parameters from requests (e.g., distance and direction for commands associated with robot movement, duration for scheduling commands) and inserted them into functions.

We constrained GPT to output responses in JSON format to support easy interpretation and execution in within our Python implementation. Overall, this combination of techniques was selected with the aim of enhancing the effectiveness and usability of the speech interface.

\subsection{Preliminary User Evaluation}

More than 10 users have trialled this interface so far subsequent to a few demonstrations. All users experienced the benefit of being able to request information from and issue commands to the robot with flexible language use. About half of them were new to the whole system and had no prior experience with robots. They could make the robot respond to some commands and requests without any knowledge of the GUI. A couple of users who tried the system prior to any demo could also do the same, without any examples of speech commands or requests to use. 

However, the users also experienced time-to-time significant delays in robot responses, primarily due to the low response speed of the OpenAI API. This issue would be amplified when used in environments with unreliable and high-latency network connections (e.g., actual disaster sites). 


\section{Robot-initiated Haptic Feedback}

Haptic feedback could convey information in scenarios where a user is overloaded with visual and/or auditory information, reducing errors and mistakes and improving performance and efficiency~\cite{gagnon2013serious}. Vibrotactile feedback that utilises vibration to simulate the skin is the most commonly used haptic feedback, as vibration motors are easily accessible, low-cost, easily powered and controlled and have small power consumption~\cite{ju2021haptic}. We designed a haptic feedback wearable using an array of vibration motors and other technologies so that a human supervisor of a human-robot team could receive important status information of a robot wirelessly, without needing to be attentive to GUI components all the time.

\subsection{Implementation of Haptic Interface}

\begin{figure*}[!htb]
\centering
\includegraphics[width=0.95\linewidth]{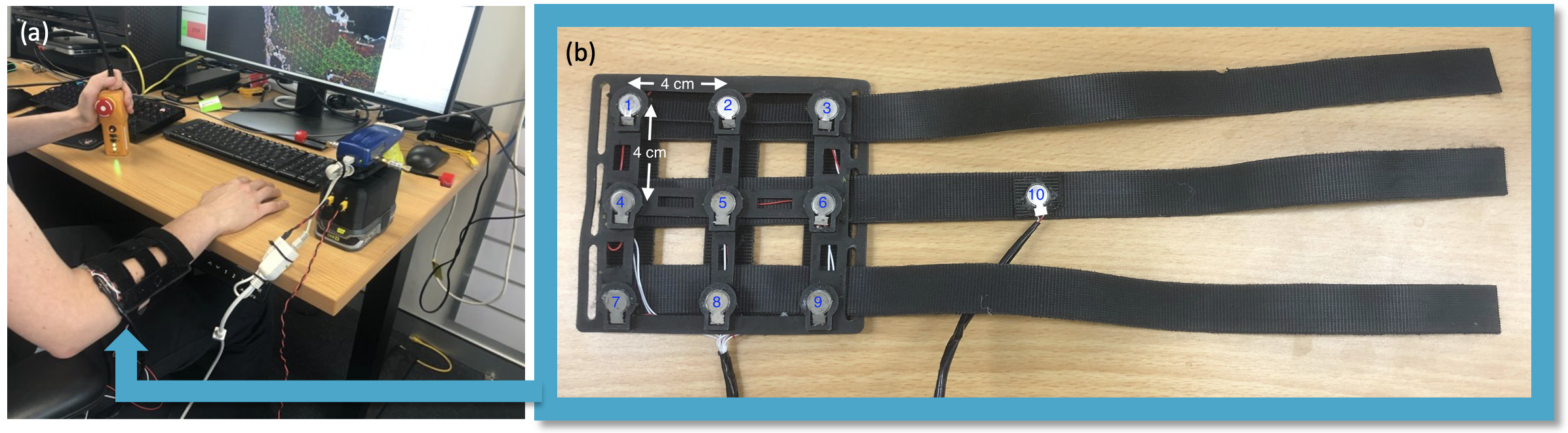}
\caption{The haptic feedback wearable (a) worn at forearm of the user while using the graphical user interface to supervise robots, and (b) a close look of motor configuration, 3D printed holder and velcro straps used to secure the wearable to the user's forearm.}
\label{fig:Haptic_wearable}
\vspace{-4mm}
\end{figure*}

\subsubsection{System Overview} The haptic wearable we designed could be worn on the human supervisor's forearm, and it consists of ten vibration motors controlled by a Raspberry Pi (see Figure 2). A Rajent node\footnote{\url{https://rajant.com/products/breadcrumb-wireless-nodes/es-series/}} has been utilised to allow the device to receive Robot status information wirelessly. We also introduced a graphical interface, so the user could configure different haptic feedback patterns and map them to different important robot status changes (see Figure 3), e.g., when an error occurs, the robot's emergency stop (EStop) is enabled or the robot being idle. 

Initially, we designed this interface to receive status information from a Spot robot, and then extended it to receive updates from up to four robots. The events being monitored by the device are given different priority levels as specified in the configuration GUI. If multiple events are detected at the same time, the event's vibration motor patterns will play out in order of priority, with lower priority events playing out while higher priority events are waiting for their realert time to elapse. If higher priority events are detected while a lower priority's pattern is playing out, the pattern will be interrupted. If multiple robots are selected for a monitored event with the same pattern, then the device will respond to the event occurring for the individual robots separately (e.g. error occurring for robot named Rat and error occurring for robot named Bingo are treated as separate events). In this case, these events are treated as being of the same priority and will not interrupt each other.

\begin{figure*}[!htb]
\centering
\includegraphics[width=\linewidth]{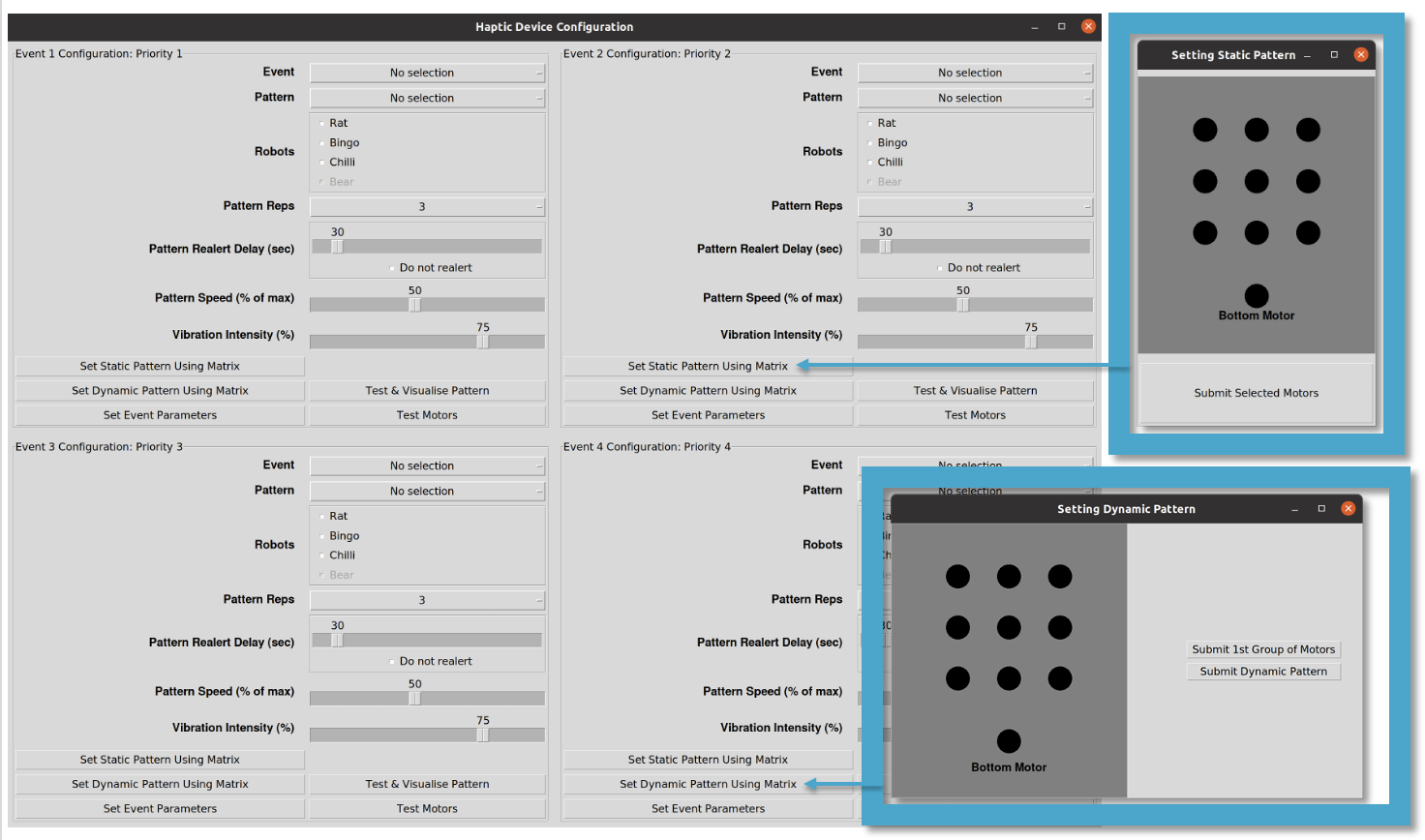}
\caption{The graphical interface for configuring haptic patterns for different status updates from multiple robots}
\label{fig:Haptic_configuration}
\vspace{-4mm}
\end{figure*}

\subsubsection{Wearable Design}

The parameters that can be altered to convey haptic information include location (where the vibration is presented), timing (when the vibration is on and when it is off), frequency of the vibration and amplitude of the vibration~\cite{erp2005waypoint}.

We selected the forearm as the body location to wear the haptic feedback device, given the high density of mechanoreceptors in the hand area, it can accommodate multiple sensors supporting diverse haptic patterns compared to other areas such as wrists, and previously haptic feedback was successfully utilised at the forearm~\cite{corniani2020tactile}. 

With the aim of creating diverse haptic patterns, we decided to use 10 motors. Since the resolution of haptic perception of the surface of the forearm has been found to be 41 mm, we designed a 3x3 matrix of vibration motors spaced 4 cm apart (centre to centre), as shown in Figure 2(a). Velcro straps are used to secure the matrix to the user's forearm. We 3D printed the holder for this vibration motor matrix using a flexible TPU (thermoplastic polyurethane) filament. After several designs and trials, we finalised this holder using minimal material to improve the user's ability to localise vibrations. The 10th vibration motor was attached to a small piece of velcro to allow for attachment to the strap underneath the user's forearm.

Since it has been found that higher frequencies (~200 Hz) are better perceived compared to lower frequencies (e.g., 20 Hz, 50 Hz, 100 Hz) at the forearm, and frequencies beyond 200 Hz would not change vibration amplitude much, therefore, achieve an equal subjective intensity rating~\cite{mahns2006vibrotactile}, we set the motors to operate at 200 Hz.

The motors were interfaced with a Raspberry Pi 3 Model B. Raspberry Pi was configured with a static IP address and connected to the Rajent node via Ethernet. Raspberry Pi was programmed with a ROS node, which received status from the robot network and controlled the motors according to the configured patterns and event priorities.

\subsubsection{Configuration Graphical User Interface} The configuration GUI, as shown in Figure 3, can be used to set parameters to monitor up to four events. For each event, the event type to be monitored can be chosen (e.g. robot idle) as well as the robots to be monitored (at least one must be chosen). The pattern type can be selected from the predefined patterns using the drop-down menu or a custom pattern can be designed using the "Set Static Pattern Using Matrix" and "Set Dynamic Pattern Using Matrix" buttons.

User could select from 3 static and 4 dynamic predefined patterns. The static patterns were: ``Pulse matrix'' (i.e., pulsing all 1-9 motors), ``Pulse bottom motor''(i.e., pulsing only the motor 10), and ``Pulse top centre motor'' (i.e., pulsing only the motor 5). The dynamic patterns were: ``Circular'' (i.e., activating and deactivating motor 4 → motor 5 → motor 6 → motor 10 one after the other), ``Moving diagonal'' (i.e., activating and deactivating motor 1 → motor 5 → motor 9 one after the other), ``Moving rows'' (i.e., activating and deactivating motors 1 + 2 + 3 → motors 4 + 5 + 6 → motors 7 + 8 + 9, one after the other), ``Moving columns'' (i.e., activating and deactivating motors 1 + 4 + 7 → motors 2 + 5 + 8 → motors 3 + 6 + 9, one after the other).

When setting a static pattern, a window will pop up, where the user can select and submit the desired motors by clicking the circles. When setting a dynamic pattern, a similar window will pop up. As dynamic patterns involve groups of motors (with the first group turned on, followed by the second, etc.),  each custom motor group can be selected using the top submit group button. Once all the desired groups of motors have been selected the dynamic pattern can be set by pressing the "Submit Dynamic Pattern" button. Note that the same event (e.g. error for Rat) cannot be mapped to more than one pattern. Event parameters have to be set in order of priority (i.e. Event 1 parameters followed by Event 2, etc.). 

The other parameters that can be set using the GUI include,
``Pattern Reps'': how many times the pattern is to repeat, ``Pattern Realert Delay'': the time between a pattern playing out and then repeating for the same event occurrence), ``Pattern Speed'': how fast the pattern plays out, and
``Vibration Intensity'': intensity of the vibration motors.
Under ``Pattern Reps", ``do not realert'' option can be used so that the pattern is only played out when the event is first detected. Once all the parameters are set, the "Set Event Parameters" button can be used to confirm selections. 

The GUI also allows the option to test the vibration motors as well as to test patterns. The ``Test Motors'' function is useful for troubleshooting as it turns each motor on and then off again one after another while printing to the terminal. The ``Test \& Visualise Pattern'' option plays out the currently selected pattern using the vibration motors while also showing a visualisation of the pattern.  

\subsection{Preliminary User Evaluation}

We tested this haptic feedback interface with 6 users, and all of them could quickly recognise robot status changes that they selected to listen to. All of them also appreciated the ability to configure patterns of their own, while highlighting the need for selecting appropriate patterns to correspond to the chosen events (e.g., a pulsing matrix pattern with high vibration intensity to signal an error and a slow pulse bottom motor pattern to signal robot idle), for ease of remembering. By not doing so, some of them struggled to identify the event type, although they detected that a status change happened in a timely manner.

\begin{table}[t]
  \centering
  \begin{tabular}{|l|c|c|c|}
   \hline
   Participant ID & P1 & P2 & P3 \\
   \hline
    Circular &	Correct	& Correct &	Correct \\
    Pulse Matrix &	Correct	& Correct	& Incorrect \\
    Pulse Bottom Motor &	Correct	& Correct	& Correct\\
    Pulse Top Centre Motor	 & Correct	& Correct	& Correct\\
    Moving Diagonal	& Correct	& Incorrect	& Correct\\
    Moving Rows	& Correct	& Correct	& Incorrect \\
    Moving Cols &	Incorrect	& Incorrect	& Incorrect \\
     \hline
  \end{tabular}
  \caption{User perception of 7 different predefined haptic patterns}
  \label{tab:1}
\end{table}

We observed the challenges that some users faced in differentially identifying certain vibration patterns when they selected to listen to multiple events. Hence, we tested the ability to differentiate 3 static patterns and 4 dynamic patterns that can be selected through the configuration interface with 3 users. Each user was introduced to these patterns without playing on the haptic device and then requested to name the randomly played patterns. Table 1 shows each of the predefined motor patterns and whether the participant was able to correctly identify the pattern. From this brief testing conducted, the patterns that appear to be best perceived are the circular, pulse bottom motor and pulse top centre motor patterns with users reporting being able to easily identify whether the top or bottom motor was vibrating but struggling to distinguish between the matrix patterns.

The users appreciated the ability to listen for multiple events while prioritising the haptic notifications for the chosen events. A few users also highlighted the usefulness of integrating the haptic feedback device with inertial measurement unit signals (IMUs) or electromyography signals (EMGs) sensors, so simple gestures can be used to ignore or stop current patterns playing out or to disable or enable all patterns, as required. Some users also suggested having a button array or small touch screen to allow for actions such as requesting information, selecting robots, and silencing alerts, as a way to change the initial configuration on the go. They also mentioned that such a display could also be used to display additional information such as the name of the robot the current event is related to when having multiple robots. Further, some users suggested adjustments to the configuration GUI, e.g., the ability to change the number of events the haptic device can monitor for (currently, 1-4 events), and enabling to open the configuration GUI during device operation to adjust parameters (currently, on startup).

\section{Future Directions}

Future work for the speech interface needs to develop solutions that can operate offline and locally aiming to minimise delays in robot response, perhaps by combining the symbolic rigour of traditional NLP pipelines with the flexibility and power of LLMs. Although we did not observe instances of hallucination, internal biases and knowledge cutoffs; common issues associated with GPT and LLMs in general~\cite{chen2023hallucination}, future work needs to investigate how these issues may present in this human-robot speech interface with further user testing and identify ways to mitigate them. Expanding this speech interface to use with a multi-robot team is another useful direction, in which we have already made some progress.

Further work for the haptic feedback interface could focus on improving its portability (e.g., by attaching the Raspberry Pi enclosure and the Rajent node pack to the user's hip). Also, future research is required for a better understanding of vibration motor pattern perception and mapping of different patterns to different events, while reducing the cognitive load of remembering.

Our plan is to design a multimodal human-robot communication system by combining the GUI, speech interface and haptic interface described in this paper, and investigate the interaction modality types suitable for communicating diverse information to human collaborators working with multiple robots to achieve optimal situational awareness. This will include running user studies with diverse participants with the use of proper assessment tools to evaluate situational awareness, cognitive load and usability.

\section{Conclusion}

The work presented in this paper involves creating interfaces for human-robot collaboration using natural language communication and haptic feedback as alternative modalities to a graphical user interface. These interfaces aim to reduce gaps in situational awareness of human team members working with robots, so they can guide robots through timely interventions to achieve missions that they cannot do alone. The preliminary user evaluations conducted provide evidence for the benefits of these alternative interfaces, as well as directions for further design and development improvements, such that the communication between human and robot members can made more natural while reducing the cognitive load of human members when working with robots. Overall, the content shared in this paper is related to two questions that the ``Empowering People in Human-Robot Collaboration'' workshop plans to address: ``Why are we creating platforms for human-robot collaboration, and what are the benefits for people?'' and ``How can the empowerment of people in HRC be facilitated through design and development?''.

\bibliographystyle{IEEEtran}
\bibliography{bibtex.bib}

\end{document}